\documentclass{ecai}

\usepackage{latexsym}
\usepackage{amssymb}
\usepackage{amsmath}
\usepackage{amsthm}
\usepackage{booktabs}
\usepackage{enumitem}
\usepackage{multirow}
\usepackage{graphicx}
\usepackage{color}
\usepackage{float}
\usepackage{placeins}
\usepackage[table,xcdraw]{xcolor}
\definecolor{mistyrose}{rgb}{1.0, 0.89, 0.88}

\newcommand{\BibTeX}{B\kern-.05em{\sc i\kern-.025em b}\kern-.08em\TeX}

\begin{document}

\begin{frontmatter}

\paperid{1405}

\title{SPIN-ODE: Stiff Physics-Informed Neural ODE for Chemical Reaction Rate Estimation}

\author[A,C]{\fnms{Wenqing}~\snm{Peng}\orcid{0009-0008-4177-7576}\thanks{Corresponding Author. Email: wenqing.peng@helsinki.fi.\\ Accepted at the European Conference on Artificial Intelligence (ECAI) 2025.}}
\author[B,C]{\fnms{Zhi-Song}~\snm{Liu}\orcid{0000-0003-4507-3097}}
\author[A,B,C]{\fnms{Michael}~\snm{Boy}\orcid{0000-0002-8107-4524}} 

\address[A]{Institute for Atmospheric and Earth Systems Research, University of Helsinki}
\address[B]{Department of Computational Engineering, Lappeenranta-Lahti University of Technology LUT}
\address[C]{Atmospheric Modelling Centre-Lahti}

\begin{abstract}
Estimating rate coefficients from complex chemical reactions is essential for advancing detailed chemistry. However, the stiffness inherent in real-world atmospheric chemistry systems poses severe challenges, leading to training instability and poor convergence, which hinder effective rate coefficient estimation using learning-based approaches. To address this, we propose a \textbf{S}tiff \textbf{P}hysics-\textbf{I}nformed \textbf{N}eural \textbf{ODE} framework (SPIN-ODE) for chemical reaction modelling. Our method introduces a three-stage optimisation process: first, a black-box neural ODE is trained to fit concentration trajectories; second, a Chemical Reaction Neural Network (CRNN) is pre-trained to learn the mapping between concentrations and their time derivatives; and third, the rate coefficients are fine-tuned by integrating with the pre-trained CRNN. Extensive experiments on both synthetic and newly proposed real-world datasets validate the effectiveness and robustness of our approach. As the first work addressing stiff neural ODE for chemical rate coefficient discovery, our study opens promising directions for integrating neural networks with detailed chemistry.
\end{abstract}

\end{frontmatter}

\section{Introduction}
\label{Introduction}

Neural Ordinary Differential Equations (Neural ODEs) have emerged as a powerful approach for representing system dynamics with neural networks. Their main advantage is the ability to model dynamics in continuous depth, offering a natural framework for time-dependent processes. They have shown significant promise in physics and chemistry~\cite{graphcast,chemode,chemode_2}. However, despite these advances, their use in detailed chemical modelling remains limited in the computer science community, particularly for tasks such as reaction rate coefficient estimation and the discovery of new chemical pathways~\cite{chemtab}.”

\begin{figure}[t]
    \centering
    \includegraphics[width=0.99\linewidth]{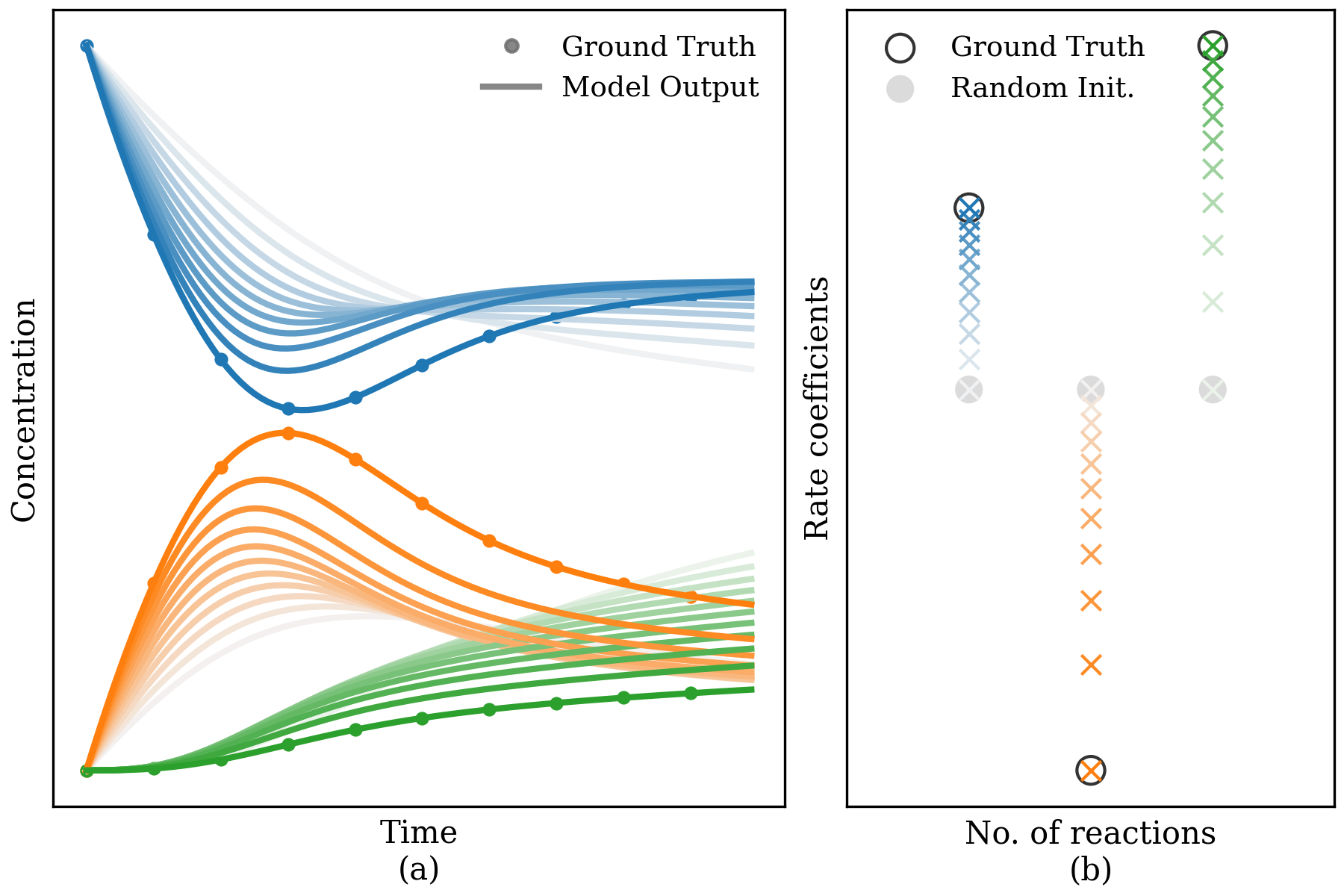}
    \caption{The proposed SPIN-ODE fits concentration trajectories (a) to infer reaction rate coefficients (b), converging from random initialisation to the true data (light colour $\rightarrow$ dark colour).}
    \label{fig:teaser}
\end{figure}

While implicit modelling approaches~\cite{air_2,air_4,air_1,chemkan,chemode} approximate concentration trajectories, explicit estimation of rate coefficients~\cite{pichelstorfer2025theory} yields mechanistic insight and can be directly applied in atmospheric research, making our target fundamentally different. This task is particularly difficult in computer science, where knowledge transfer from domain-specific numerical chemistry to neural networks is still at an early stage. The fundamental obstacle lies in identifying and encoding the essential factors, such as temperature dependencies and chemical reaction paradigms, into a learnable and generalisable model architecture.

Moreover, the scarcity of real experimental data further complicates this endeavour. Most available measurements come from highly controlled environments like flow-tube~\cite{flow_tube} or smog-chamber~\cite{smog} experiments, which are costly, time-consuming, and often prone to significant measurement uncertainties. As a result, purely data-driven learning approaches suffer from limited generalisation and a lack of physical grounding. Furthermore, chemical reaction systems are often inherently stiff, characterised by vastly different reaction timescales, which significantly complicates numerical integration and destabilises neural network training. To address this, our method incorporates domain-informed knowledge derived from detailed numerical chemical models, allowing the neural network to learn subject to physical constraints and chemical principles even in data-sparse and numerically stiff regimes.

In this work, we propose a novel framework that leverages the strengths of neural ODE to estimate reaction rate coefficients. As illustrated in Figure~\ref{fig:teaser}, our goal is to uncover the correlations between chemical concentration trajectories and rate coefficients, so that the optimal rate coefficients can be found by ODE fitting. Rather than using neural networks as an implicit ODE system, our goal is to discover the hidden rate coefficients for detailed chemistry. We envision this integration of AI and detailed chemistry for scientific discovery, potentially uncovering new patterns, simplifying mechanisms, and offering interpretable insights into complex reaction systems. To summarise:
\begin{itemize}
    \item We propose an explicit stiff physics-informed neural ODE framework (SPIN-ODE) for detailed chemistry. This framework efficiently models ODE trajectories while simultaneously extracting interpretable reaction rate coefficients given reaction pathways.
    \item We introduce a three-stage optimisation strategy specifically designed to handle stiffness and enhance training stability. Our novel physical loss functions and trajectory resampling techniques significantly improve ODE fitting and rate constant estimation.
    \item Compared to existing approaches that primarily calibrate rate coefficients of stiff reactions based on close initial guesses, our method is the first work to automatically find optimal rate coefficients in the absence of prior knowledge.
\end{itemize}

\section{Related Work}
\label{related_work}
Our study addresses detailed chemical kinetics with neural networks, building on advances in neural ODEs and their applications in chemical modelling.

\subsection{Numerical and Neural ODEs}
Numerical methods for solving ordinary differential equations have been extensively studied, with classical approaches such as Euler’s method, Runge-Kutta schemes, and multistep methods~\cite{rk} forming the foundation of the field. These techniques aim to approximate solutions to initial value problems with varying degrees of accuracy, stability, and computational efficiency. Over the decades, significant advances have been made in adaptive step-size control, stiff equation solvers, and error estimation strategies. Recent developments have also explored the intersection of numerical analysis and machine learning, where neural networks and physics-informed approaches offer data-driven alternatives for solving ODEs. Neural ODE was first introduced in~\cite{ODE}, where backpropagation can be used to train an ODE model. Later on, many works~\cite{ode_1,ode_2,ode_3,ode_4,augmented,steer} followed up on this direction to further analyse and improve the performance. For example, \cite{ode_2} analyses the theory behind the infinite-depth neural ODEs and proposes adaptive-depth and data-controlled neural ODE paradigms. \cite{augmented} demonstrates that a simple augmentation of the latent ODE space can achieve lower losses with few parameters and is more stable to train. To overcome the slow training process, \cite{ode_1} proposes a regularised neural ODE (RNODE) to learn well-behaved dynamics. ~\cite{ode_6} proposes Fourier analysis to ODEs for temporal estimation and potentially high-order spatial gradients calculation. Later on, neural ODEs are also applied to domain-specific applications, like time-series processing~\cite{ode_3}, latent space image processing~\cite{ode_4}, glucose dynamics~\cite{ode_5} and robotics control~\cite{ode_7}.

Meanwhile, a significant challenge to the current neural ODEs is the stiff ODE system. \cite{stiff_1} was the first to observe that explicit methods fail for certain chemical kinetic systems. Even when the observed concentration trajectories appear smooth, stiffness arises because some reactions evolve extremely fast while others are much slower, leading to hidden oscillatory dynamics and numerical instability. Kim et al.~\cite{kim_stiff_2021} propose stiff neural ODE, which introduces scaling and an adjoint solution strategy to mitigate stiffness, and demonstrate its effectiveness on two classical stiff chemical systems. Fronk et al. present two consecutive studies~\cite{stiff_3,stiff_4} to demonstrate both explicit and implicit solution to stiff neural ODE training. Physics-informed neural network (PINN)~\cite{pinn} is another solution by integrating more physical constraints to tackle the high-stiff regime. Ji et al.~\cite{ji_stiff-pinn_2021} develop Stiff-PINN, which integrates Quasi-Steady-State-Assumptions (QSSA) into the network by predicting only non-QSS species, reconstructing QSS species algebraically, and constraining the physics-informed loss with the reduced system, thereby mitigating stiffness in chemical kinetics. Seiler et al.~\cite{stiff_5} present a method based on transfer learning and design a multi-head PINN network to divide and conquer the ODE system into low-stiff and high-stiff regions. Sulzer et al.~\cite{stiff_6} propose an autoencoder to solve stiff ODE in the latent space. The first- and second-order gradients are used as physical losses for optimisation.

\subsection{Chemical Neural Networks}
The rise of deep learning in the early 2010s has significantly expanded the scope of scientific discovery processes. Neural networks for chemical modelling have recently attracted significant research interest, like weather forecasting~\cite{graphcast,pangu,aurora}, air quality estimation~\cite{air_1,air_2,air_3,air_4}, and others. Chemical kinetics can be considered as a complex ODE system and researchers have been studying both implicit and explicit approaches for efficient chemical modelling. For example, Owoyele and Pal~\cite{chemode} utilise neural ODE to simulate complete chemical kinetics implicitly for fast computation. Goswami et al.~\cite{goswami2024learning} leverages the recently proposed neural operator~\cite{lu2021learning} to exploit the dynamical structure of chemical reactions. Liu et al.~\cite{chemode_2} further improve it by proposing ChemODE, an attention based neural operator for fast and accurate chemical modelling, which can successfully estimate over hundreds of chemical reactions. Kolmogorov-Arnold Networks (KAN) were presented recently~\cite{kan} as an alternative to multi-layer perceptrons for general neural network applications. It has attracted significant attention for chemical modelling~\cite{kanode}. ChemKAN~\cite{chemkan} was later introduced to learn hydrogen combustion chemistry modelling, achieving a twofold speed-up over detailed numerical chemistry.

In contrast to implicit models that approximate the overall chemical system behaviour, explicit chemical neural networks aim to directly learn or represent the fine-grained chemical mechanisms themselves, such as activation energies, rate coefficients or reaction pathways. This explicit modelling is crucial for advancing scientific understanding, especially in domains where interpretability and physical consistency are as important as predictive performance. Despite its importance, relatively few studies have focused on using neural networks to capture detailed chemical kinetics in an interpretable and physically faithful manner. This gap is largely due to the complexity of chemical reaction networks and the challenge of integrating domain knowledge, such as thermodynamic constraints and reaction mechanisms, into machine learning models. Some notable efforts include ChemTab~\cite{chemtab}, which tabulates reaction rates using neural networks for combustion modelling, and ReactionNet~\cite{reactionnet}, which predicts reaction rates by incorporating reaction templates and thermochemical data. Ji et al.~\cite{ji_autonomous_2021} autonomously discover reaction pathways and rates by estimating stoichiometric matrix and rate coefficients. However, it is only used to calibrate the rate coefficients initialised with a small-scale uncertainty given fixed reaction pathways~\cite{su_kinetics_2023}. In this work, we argue that explicit neural representations of chemical reactions offer a promising direction for coupling accuracy, interpretability, and computational speed in scientific modelling, especially when integrated with recent developments in symbolic neural networks or physics-informed learning frameworks.

\section{Approach}
\label{approach}

\begin{figure*}[!t]
	\centering
		\centerline{\includegraphics[width=2\columnwidth]{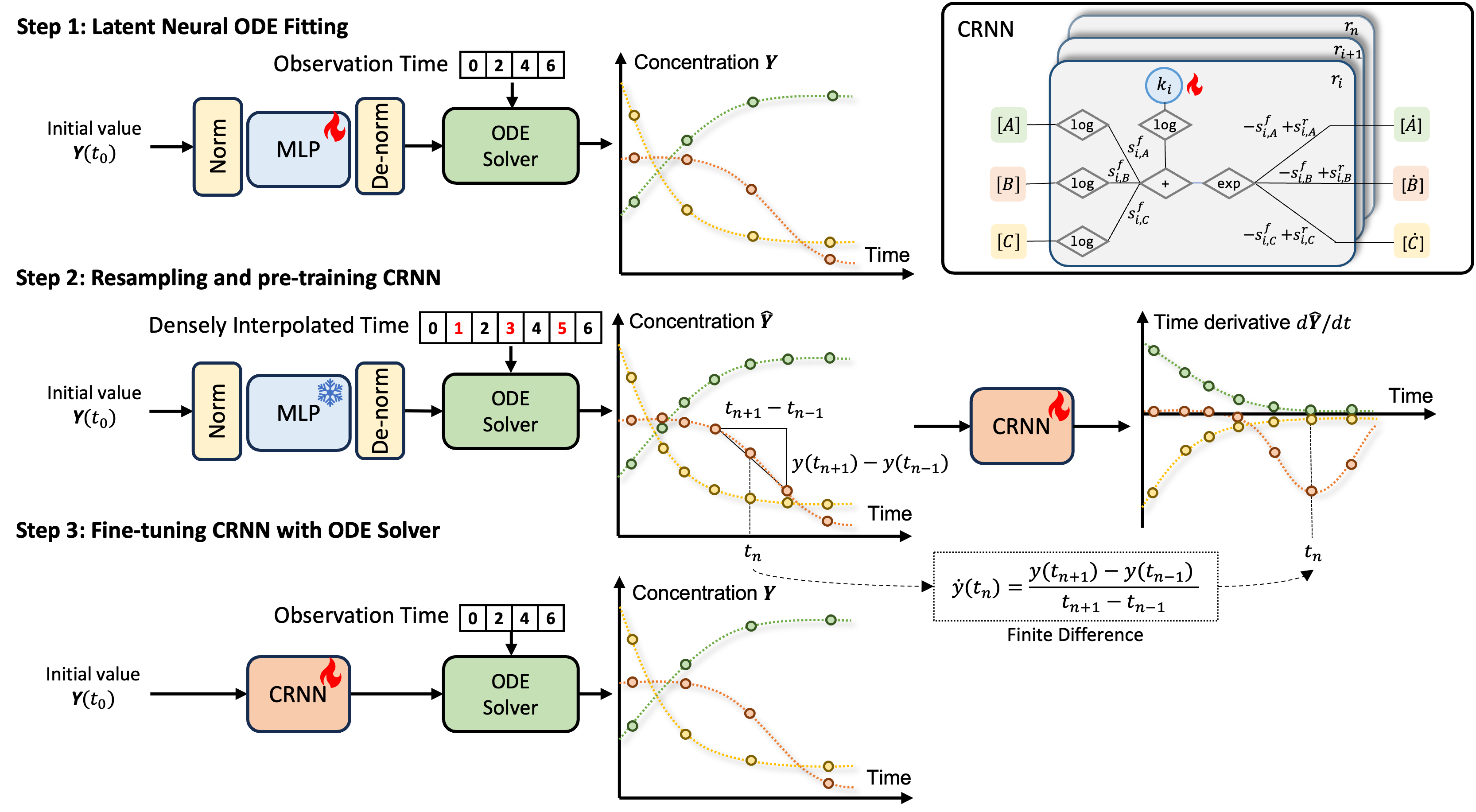}}
		\caption{\small{\textbf{Our proposed SPIN-ODE framework for stiff reaction rate coefficient estimation.} It includes three steps: 1) black-box neural ODE fitting by optimising predicted ODE trajectories, 2) resampling data from step 1 and pre-training physics-based CRNN by optimising predicted time derivatives, and 3) fine-tuning CRNN by optimising predicted concentration trajectories. The detailed structure of the CRNN is shown in the upper right corner.
		}}
		\label{fig:network}
\end{figure*}

\subsection{Preliminaries} 
\noindent \textbf{1. Chemical kinetics and stiff ODEs.} We illustrate the chemical kinetics with the following example Robertson~\cite{robertson1966numerical} reaction system:
\begin{equation} \label{react: rober}
    \begin{aligned}
    \textrm{A} &\rightarrow \textrm{B}, \\
    \textrm{B} + \textrm{B} &\rightarrow \textrm{C} + \textrm{B},  \\
    \textrm{B} + \textrm{C} &\rightarrow \textrm{A} + \textrm{C}.
    \end{aligned}
\end{equation}
Without loss of generality, any of these reactions can be represented in a unified form as:
\begin{equation} \label{eq:reaction}
    s_{i,a}^f\textrm{A} + s_{i,b}^f\textrm{B} + s_{i,c}^f\textrm{C}\rightarrow s_{i,a}^r\textrm{A} + s_{i,b}^r\textrm{B} + s_{i,c}^r\textrm{C},
\end{equation}
in which $s_{i,j}^f$ and $s_{i,j}^r$ are the forward and reverse stoichiometric coefficients for species $j$ in reaction $i$, indicating molecules consumed and produced per reaction. The reaction rates are determined by a power-law expression involving the rate coefficients $k_i$ and reactant concentrations:
\begin{equation} \label{eq:rate}
    r_i = k_i[A]^{s_{i,a}^f}[B]^{s_{i,b}^f}[C]^{s_{i,c}^f}.
\end{equation}
Here we use $[\cdot]$ to represent chemical concentrations. Reaction rate coefficients $k_i$ follows the Arrhenius relationship and is dependent on the temperature and molecular energy barriers, and is often determined experimentally. The overall rate of change for species concentration $Y_j$ is the sum of the consumption and production rates from all reactions:
\begin{equation}
    \frac{d[Y_j]}{dt} = \sum_{i} -s_{i,j}^f r_i + s_{i,j}^r r_i.
\end{equation}
Therefore, the governing equations for the Robertson reaction system, shown in Eq.~\eqref{react: rober}, constitute a system of non-linear ODEs:
\begin{equation} \label{eq:rober}
\begin{aligned}
    \frac{d[A]}{dt} &= -k_1[A] + k_3[B][C],\\
    \frac{d[B]}{dt} &= k_1[A] - k_2[B]^2 - k_3[B][C],\\
    \frac{d[C]}{dt} &= k_2[B]^2.
\end{aligned}
\end{equation}

When reaction timescales vary significantly, solving the ODE with ordinary numerical methods becomes unstable, and the system is identified as \textbf{numerically stiff}. The stiffness of a system is often characterised by the stiffness ratio, defined as the disparity between the maximum and minimum eigenvalues $\lambda$ of the system Jacobian:
\begin{equation}
    \text{Stiffness ratio} = \frac{\max|Re(\lambda)|}{\min|Re(\lambda)|}.
\end{equation}
High stiffness ratios are typical in atmospheric chemistry. Notably, in cases like the classic stiff Robertson problem, zero eigenvalues result in the stiffness ratio being undefined. While numerical stiff ODE solvers are extensively studied, their robust integration into machine learning frameworks remains limited.

\noindent\textbf{2. Neural ODE.} Similar to chemical reactions, many systems across science and engineering domains can be represented by ODEs:
\begin{equation}
    \frac{dy(t)}{dt} = f(y(t), t),
\end{equation}
where $y(t)$ is the system state variable at time $t$, and the function $f$ governs the system dynamic. Neural ODE approximates unknown system dynamics $f$ using neural networks parametrised by $\theta$:
\begin{equation}\label{eq:neural_diff_eq}
    \frac{dy(t)}{dt} = NN_\theta(y(t), t).
\end{equation}
Since observational data typically includes only discrete-time state variables $y(t)$, neural ODE integrates the approximated dynamics to predict system states at subsequent times:
\begin{equation} \label{eq:odesolve}
    \begin{aligned}
        y(t_1) &= y(t_0) + \int_{t_0}^{t_1}NN_\theta(y(t), t)dt\\
               &= ODESolver(NN_\theta, y(t_0), t_0, t_1).
    \end{aligned}
\end{equation}
The loss function is defined as the distance, e.g., mean squared error (MSE), between model predictions and observed state variables.
With the differentiable $ODESolver$, the gradient of the loss with respect to the network parameters $\theta$ can be calculated and leveraged to optimise the network using gradient-based algorithms. Chen et al. propose a widely-used approach for memory-efficient gradient calculation by solving the backward-in-time~\cite{ODE}. However, neural ODE training is extremely vulnerable to stiffness, where minor errors in the forward pass are rapidly amplified in the backward pass~\cite{kim_stiff_2021}. In this paper, we are specifically interested in solving stiff ODEs in atmospheric chemistry for rate coefficient estimation. Existing stiff neural ODE solutions, like collocation based ODE training~\cite{huang_training_2025,roesch_collocation_2021} and temporal regularisation based ODE training~\cite{steer} fail to fit ODE trajectories, and cannot extract the rate coefficients.

\subsection{Proposed method} \label{sec:our_approach}
We propose a novel neural network framework, SPIN-ODE, for stiff chemical modelling, which not only learns ODE trajectories but also explicitly extracts reaction rate coefficients for detailed chemistry. As illustrated in Figure~\ref{fig:network}, the framework addresses stiffness through a three-step incremental learning strategy that combines the robustness of black-box stiff neural ODE with the physical interpretability of CRNN. First, a neural ODE is trained to fit concentration trajectories from initial chemical conditions. Second, the learned trajectories are resampled to pre-train the CRNN for rate coefficient estimation. Finally, the pre-trained CRNN is integrated with a stiff ODE solver to fine-tune the rate coefficients.

\noindent \textbf{SPIN-ODE step 1: Trajectory fitting with a black-box neural ODE}  
In the first step, we fit the trajectories using a black-box stiff neural ODE model, where the dynamics are learned by a simple multilayer perceptron (MLP) combined with an ODE solver. After training, the MLP approximates the mapping between $y$ and $dy/dt$. A key component is the design of the normalisation layers, which are essential for stable ODE fitting in stiff chemical systems. Our motivation follows ~\cite{kim_stiff_2021}, which shows that stiffness arises from scale separations in concentrations and their time derivatives, and can be mitigated by scaling both to comparable ranges for stable optimisation. Specifically, we describe the process as follows:
\begin{equation}
    \frac{dy(t)}{dt} = NN_\theta(\frac{y(t)-y_\text{min}}{y_\text{max} - y_\text{min}}, t)\frac{y_\text{max} - y_\text{min}}{t_\text{scale}},
\end{equation}
where $y_\text{max}$ and $y_\text{min}$ are the species-wise maximum and minimum concentrations used in the normalisation layer, and $t_\text{scale}$ is the time span of the entire measurements. The ODE solver integrates the output time derivative to a trajectory as Eq.~\eqref{eq:odesolve}, and the loss function is correspondingly normalised by the scale of concentrations as:

\begin{equation} \label{eq:scaleMSE}
    \mathcal{L}_y = \text{MSE}(\frac{y(t)^\text{model}}{y_\text{max} - y_\text{min}}, \frac{y(t)^\text{observe}}{y_\text{max} - y_\text{min}}).
\end{equation}

In addition to the loss on predicted state variables, we introduce two derivative-based loss terms on the velocity and acceleration of concentration changes. These act as a stabilisation strategy to regularise the training under stiffness, rather than as strict physical constraints. Specifically, we compute the 1st-order and 2nd-order derivatives losses ($\mathcal{L}_{\dot{y}}$ and $\mathcal{L}_{\ddot{y}}$) with respect to time, then apply the same scaling technique as in Eq. \eqref{eq:scaleMSE}, and the final loss function is:

\begin{equation} \label{eq:total_loss}
    \mathcal{L} = \mathcal{L}_y + \alpha \mathcal{L}_{\dot{y}} + \beta \mathcal{L}_{\ddot{y}},
\end{equation}
where $\alpha$ and $\beta$ are both set to $0.1$.

\noindent \textbf{SPIN-ODE step 2: Pre-training CRNN with estimated derivatives.}
While non-linear system dynamics can be approximated using simple feed-forward networks such as MLPs, the parameters learned by such black-box models typically lack interpretability. In contrast, physics-based neural ODE encodes parameters and mathematical operators derived directly from physical equations~\cite{ji_autonomous_2021, kong_dynamic_2022}. Following~\cite{ji_autonomous_2021, su_kinetics_2023}, we leverage the Chemical Reaction Neural Network (CRNN) (see Figure~\ref{fig:network}), a physics-based neural network tailored for discovering chemical reaction parameters. Given fixed reaction pathways, CRNN reformulates the rate law in Eq.~\eqref{eq:reaction} as a linear combination of the logarithm of learnable rate coefficients $\theta_i$ and the logarithms of species concentrations, followed by an exponential transformation:
\begin{equation}\label{eq:log_rate_law}
    r_i = \exp\left(\ln(\theta_i) + s_{i,a}^f \ln[A] + s_{i,b}^f \ln[B] + s_{i,c}^f \ln[C]\right).
\end{equation}
Each CRNN node corresponds to an individual reaction, and nodes can be stacked to represent a reaction system. The overall time derivative of species concentrations $d\mathbf{Y}/dt$, is obtained by linearly transforming the reaction rate vector $\mathbf{R}$, through the net stoichiometric matrix $\mathbf{S}$, defined as $s_{i,j} = -s_{i,j}^f + s_{i,j}^r$:
\begin{equation}
    \frac{d\mathbf{Y}}{dt} = \mathbf{S} \cdot \mathbf{R}.
\end{equation}

To facilitate CRNN training, we deviate from prior work in which the CRNN output is directly integrated and trained end-to-end on concentration trajectories. Such coupling is prone to instability in stiff regimes, since small integration errors can rapidly accumulate during back-propagation. Instead, in SPIN-ODE we decouple the CRNN from ODE integration and train it as a supervised model of the differential equations as in Eq.~\eqref{eq:neural_diff_eq}. Specifically, we first generate paired data of concentration and time derivative from the fitted trajectories in step 1: when the black-box neural ODE fits the concentration trajectory ${ {y}(t_i) }_{i=1}^M$, we increase the temporal resolution by interpolation to obtain ${ \hat{y}(t_i) }_{i=1}^N$ with $N > M$. We then approximate the derivatives $\dot{\hat{y}}(t)$ at state $\hat{y}(t)$ using finite difference:

\begin{equation} \label{eq:diff}
    \dot{\hat{y}}(t_i) \approx 
    \begin{cases}
        \dfrac{\hat{y}(t_{i+1}) - \hat{y}(t_{i-1})}{t_{i+1} - t_{i-1}}, & 1 < i < N , \\
        \dfrac{\hat{y}(t_{2}) - \hat{y}(t_{1})}{t_{2} - t_{1}}, & i = 1 , \\
        \dfrac{\hat{y}(t_{N}) - \hat{y}(t_{N-1})}{t_{N} - t_{N-1}}, & i = N .
    \end{cases}
\end{equation}

\begin{table*}[!t]
\centering
\renewcommand\arraystretch{1.3}
\resizebox{\textwidth}{!}{
\begin{tabular}{c|cclcclcc}
\toprule
Dataset  & \multicolumn{2}{c}{Robertson}        &  & \multicolumn{2}{c}{POLLU}        &  & \multicolumn{2}{c}{AOXID}        \\ \midrule
Task     & ODE fitting       & Rate coeff. estimate &  & ODE fitting       & Rate coeff. estimate &  & ODE fitting       & Rate coeff. estimate \\ \midrule
Ji et al.~\cite{ji_autonomous_2021} & $2.2\times10^{-2}$ &    1.93&  & $9.3\times10^{19}$ &             9.25&  & *                  &             *\\
\cellcolor{mistyrose}{SPIN-ODE-s1} & \cellcolor{mistyrose}{$2.8\times10^{-3}$} &   \cellcolor{mistyrose}{-}& \cellcolor{mistyrose}{} & \cellcolor{mistyrose}{$1.1\times10^{-5}$} &  \cellcolor{mistyrose}{-}& \cellcolor{mistyrose}{} & \cellcolor{mistyrose}{$1.4\times10^{-5}$} &  \cellcolor{mistyrose}{-}\\
\cellcolor{mistyrose}{SPIN-ODE-s2} & \cellcolor{mistyrose}{$3.4\times10^{-3}$} &             \cellcolor{mistyrose}{0.69}& \cellcolor{mistyrose}{} & \cellcolor{mistyrose}{$1.9\times10^{-2}$}&   \cellcolor{mistyrose}{1.42} & \cellcolor{mistyrose}{} & \cellcolor{mistyrose}{$8.9\times10^{-5}$}&   \cellcolor{mistyrose}{0.15}\\
\cellcolor{mistyrose}{SPIN-ODE-s3} & \cellcolor{mistyrose}{\textbf{$6.3\times10^{-6}$}} &             \cellcolor{mistyrose}{\textbf{$7.4\times10^{-3}$}} & \cellcolor{mistyrose}{} & \cellcolor{mistyrose}{\textbf{$8.6\times10^{-6}$}} &             \cellcolor{mistyrose}{0.46} & \cellcolor{mistyrose}{} & \cellcolor{mistyrose}{\textbf{$4.5\times10^{-6}$}} &             \cellcolor{mistyrose}{0.10}\\ \bottomrule
\end{tabular}
}
\caption{\textbf{Overall comparisons between our proposed three-step SPIN-ODE framework and Ji et al.~\cite{ji_autonomous_2021} model.} We show the scaled MSE loss of the ODE trajectory fitting, as well as the mean absolute error on log-scaled rate coefficient estimation. - means the network involved in the current step does not provide estimation, while * means runtime error.}
\label{tab:loss_y}
\end{table*}

This design offers two main benefits. First, by training CRNN directly on interpolated and thus smoothed $(\hat{y}, \dot{\hat{y}})$ pairs rather than coupling it with an ODE solver, we avoid unstable integration during the early stage of CRNN training. Once the parameters are reasonably initialised, ODE integration can be safely reintroduced in step 3 for fine-tuning. Second, keeping the CRNN parameters aligned with physical rate laws preserves interpretability, which would be compromised if the scaling techniques from step 1 were applied directly. In that case, reaction rates would be non-linearly distorted, and the learned coefficients would lose their physical meaning.

\noindent \textbf{SPIN-ODE step 3: Fine-tuning CRNN with ODE integration.}
After pre-training, the CRNN is coupled with a stiff ODE solver (Eq.~\eqref{eq:odesolve}) to further refine the rate coefficients. Unlike step 2, where supervision is limited to discrete $(\hat{y}(t_i), \dot{\hat{y}}(t_i))$ pairs, the solver integrates through intermediate time steps, thereby constraining the dynamics between sampled points and enforcing consistency across the full trajectory. Eventually, the optimised parameters $\theta_i$ from CRNN are taken as the estimated reaction rate coefficients $k_i$.

\section{Experiments}
\label{experiment}

\begin{figure*}[t!]
	\centering
            \centerline{\includegraphics[width=2\columnwidth]{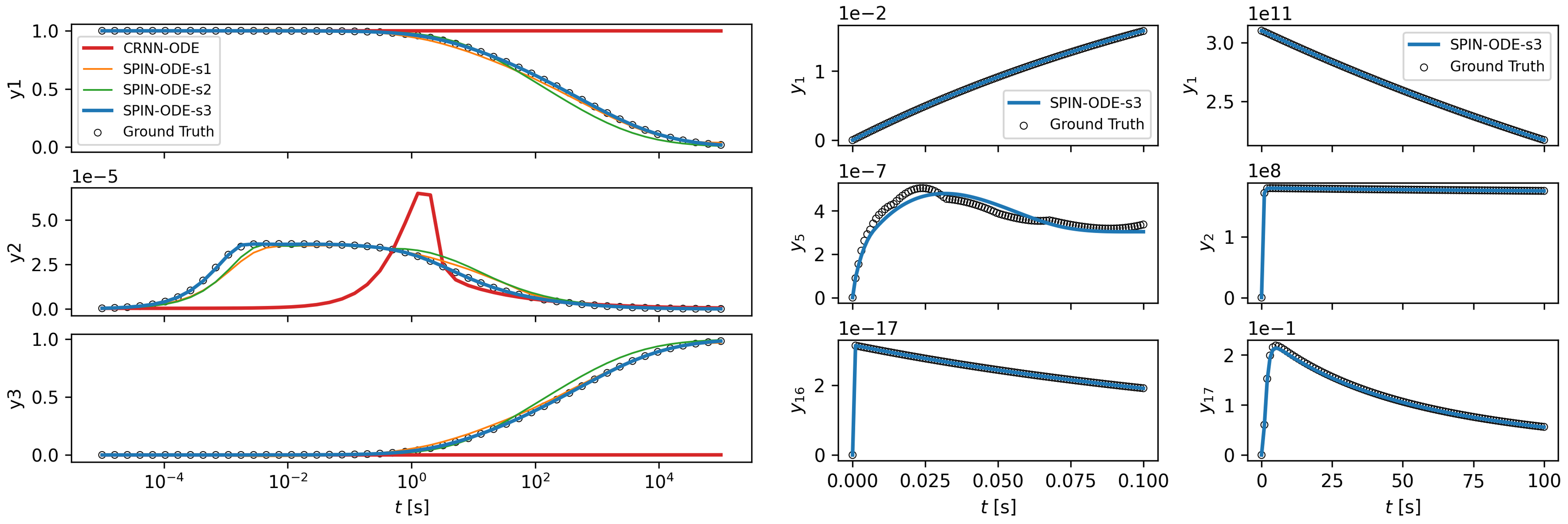}}
		\caption{\small{\textbf{Concentration trajectory fitting compared with observation.}
        From left to right are the results for Robertson, POLLU and AOXID problems. 3 representative concentration trajectories are selected from the POLLU and AOXID problems. Results for Robertson show a gradual approximation to ground truth with our incremental approach.
        }}
		\label{fig:fit_y}
\end{figure*}

\begin{figure*}[!t]
	\centering
            \centerline{\includegraphics[width=2\columnwidth]{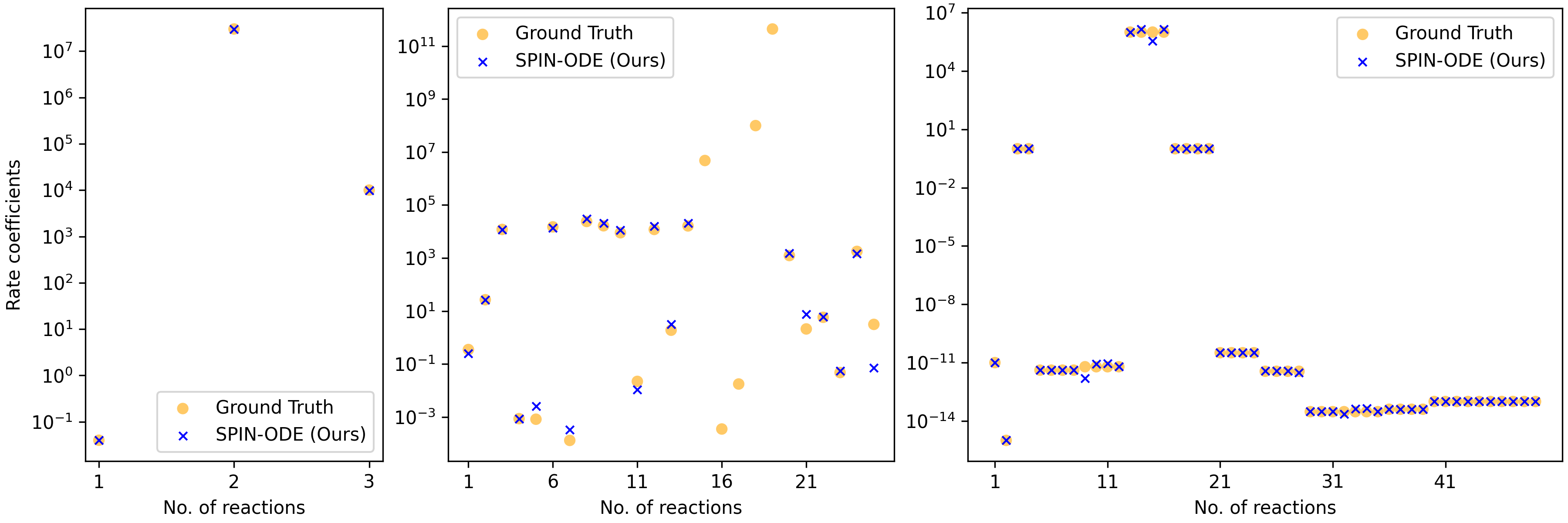}}
		\caption{\small{\textbf{Estimated rate coefficients for Robertson, POLLU and AOXID, from left to right.} The blue cross symbol represents the model prediction, and the yellow circle represents the ground truth value.
        }}
		\label{fig:rate}
\end{figure*}

\subsection{Experimental setups}
Our method is first validated using the \textbf{Robertson} problem, a classical benchmark comprising three chemical species and three stiff reactions as presented in Eq.~\eqref{react: rober}. The governing equations for the Robertson problem are explicitly defined in Eq.~\eqref{eq:rober}, with ground-truth reaction rate coefficients as $[0.04, 3\times10^7, 1\times10^4]$. 

Real-world atmospheric chemistry systems typically involve significantly more complex dynamics, exhibiting substantial stiffness across numerous reactions. Consequently, we further evaluate our method on the \textbf{POLLU} problem~\cite{pollu}, a widely utilised atmospheric chemistry model for stiff ODE benchmarks. This system includes 25 reactions among 20 chemical species, with reaction rate coefficients spanning from $10^{-3}$ to $10^{12}$. In addition, we include an autoxidation reaction (\textbf{AOXID}) scheme involving 44 species and 49 reactions. It is a simplified system representing realistic autoxidation processes in the atmosphere characterised by intricate radical chain mechanisms~\cite{pichelstorfer2025theory}. The reaction rate coefficients in this system vary dramatically, ranging from approximately $10^{-14}$ to $10^{6}$. Comprehensive details for the POLLU and AOXID reaction schemes are provided in the supplementary materials~\cite{peng2025spin}.

Datasets for the Robertson and POLLU problems are numerically generated using a stiff-aware Backward Differentiation Formula (BDF) solver available in the SciPy library. The AOXID dataset is generated using the Rosenbrock solver within the well-known Kinetic PreProcessor (KPP)~\cite{damian_kinetic_2002}. For the Robertson system, initial concentrations are set as $[y_1, y_2, y_3]=[1,0,0]$, with 50 data points sampled logarithmically spaced between $10^{-5}$ and $10^{5}$ seconds. For the POLLU and AOXID datasets, initial conditions detailed in the supplementary material are used, and 100 data points are uniformly sampled across the time span of 0.1 and 100 seconds, respectively. To mitigate numerical instabilities arising from prolonged integration and accumulation of errors, a sliding window approach of size 20 and stride of 10 is adopted during the training for POLLU and AOXID.

The proposed SPIN-ODE leverages the stiff-aware differentiable ODE solver \texttt{Kvaerno3} from the \texttt{diffrax} library \cite{kidger_neural_2022}. %
For the POLLU dataset, the MLP used in step 1 has an input and output dimension of 20 and a hidden layer size of 128, resulting in a total of 21,780 trainable parameters. In contrast, the CRNN in steps 2 and 3 has only 25 trainable parameters, corresponding to the number of reactions in the system. Network parameters are optimised using the Adam algorithm, initialised with a learning rate of 0.001, which is annealed when training loss does not decrease for 10\% of the total epochs. All experiments were conducted on a single GPU node of the CSC Mahti supercomputer, using one NVIDIA A100-SXM4-40GB GPU and 16 cores of an AMD Rome 7H12 CPU. For the POLLU dataset, the runtimes for the three stages of SPIN-ODE were 1.17 h, 6.26 min, and 11.83 min, respectively. Convergence curves for all three training stages are shown in Figure 1 of the supplementary material~\cite{peng2025spin}. The code is available at \textcolor{magenta}{\texttt{https://github.com/pvvq/SPIN-ODE}}, with an archived version in~\cite{peng_spinode_2025}.

\subsection{Comparison with state of the art}
Our overall comparisons include both ODE trajectory fitting and rate coefficient estimation. Given the wide-ranging magnitudes of the chemical concentration and reaction rate coefficients, we evaluate the model estimation errors on a normalised scale. To the best of our knowledge, this is the first study to address stiff chemical ODEs for rate coefficient estimation. The only closely related work is by Ji et al.~\cite{ji_autonomous_2021}, which we reimplemented for comparison.

The quantitative results on ODE fitting and rate coefficients are reported in Table~\ref{tab:loss_y}. We show our results in every step (SPIN-ODE-s1, SPIN-ODE-s2, SPIN-ODE-s3).
As shown in Table~\ref{tab:loss_y}, ours outperforms Ji et al.~\cite{ji_autonomous_2021} in every step. We note that Ji et al. fail to converge on the AOXID dataset, and a large error on the POLLU dataset. On the other hand, our final results (SPIN-ODE-s3) achieve low errors in rate coefficient estimation and ODE fitting, specifically $7.4\times10^{-3}$ for Robertson, $8.6\times10^{-6}$ for POLLU, and $4.5\times10^{-6}$ for AOXID. The promising results on the AOXID dataset indicate that our approach could be used in real-world atmospheric chemistry. Furthermore, by comparing SPIN-ODE-s3 with SPIN-ODE-s2, it demonstrates the benefit from the final refinement process. Note that in SPIN-ODE-s1, we use MLPs to implicitly learn ODE fitting and it cannot estimate rate coefficients yet. Compared to SPIN-ODE-s1, SPIN-ODE-s2 exhibits larger ODE fitting errors. These errors result from the oscillatory nature of derivatives in stiff systems, and training on smoothed derivatives captures only the slower components rather than the full system dynamics. This further indicates the necessity of the third fine-tuning step.

To better visualise the ODE trajectory fitting, we plot the trajectory predictions in Figure~\ref{fig:fit_y}. From left to right, we have the results for Robertson, POLLU and AOXID. Note that we select the three most representative reactions from POLLU and AOXID for visualisation. The complete reaction plots can be found in the supplementary material. Since the method by Ji et al. performs reliably only on the Robertson problem, we did not show its trajectories on POLLY and AOXID. For our approach, it fits the ground truth trajectories in all three cases. 

Since our primary focus is on rate coefficient estimation, we also plot the scatter diagram to demonstrate all rate coefficients in Figure~\ref{fig:rate}. We can see that the model predictions (blue cross) closely match with ground truth values (yellow circles). Note that for the POLLU problem, coefficients 15-19 correspond to hydroxyl (OH) cycling reactions, whose rate coefficients have been intensively studied and are provided during training to facilitate convergence, and thus excluded from the estimation result.

\subsection{Ablation studies}
We compare three alternative strategies against our proposed SPIN-ODE-s2 with interpolation to highlight its design choices: (1) applying finite differences directly to the original observations to obtain time derivatives, without SPIN-ODE-s1 MLP fitting (referred to as \textit{Direct FD}); (2) using the MLP outputs from SPIN-ODE-s1 to train the CRNN, without resampling or differentiating (referred to as \textit{MLP proxy}); and (3) applying finite differences to resampled trajectories as in SPIN-ODE-s2, but without interpolation. To ensure fairness, all methods without interpolation are trained for proportionally more epochs to match the number of collocation points.

\begin{table}[t]
\centering
\renewcommand\arraystretch{1.3}
\resizebox{\columnwidth}{!}{
\begin{tabular}{c|cc}
\toprule
                             & ODE fitting    & Rate coeff. estimate \\ \midrule
Direct FD                   & $3.4\times10^{-4}$ & 0.192           \\
MLP-proxy& $1.1\times10^{-6}$ & 0.167\\
SPIN-ODE-s2 w/o Interpolation&                    $7.4\times10^{-5}$&                 0.173\\
SPIN-ODE-s2 w/ Interpolation& $8.9\times10^{-5}$& 0.145\\ \bottomrule
\end{tabular}
}
\caption{\textbf{Ablation studies.} We report the quantitative results for our pre-training step on the AOXID dataset with or without key proposed components.}
\label{tab:coll_diff}
\end{table}

Table~\ref{tab:coll_diff} presents the results on the AOXID dataset. Applying finite differences directly to sparse observations performs worst in both trajectory prediction and parameter estimation, reflecting strong sensitivity to data resolution. Using the black-box MLP output as a proxy for derivatives yields the best trajectory match ($\mathcal{L}_y = 1.1 \times 10^{-6}$), but its higher rate coefficient error shows that reproducing trajectories alone does not guarantee reliable parameter estimation, consistent with our earlier discussion of the disparity between system dynamics revealed by concentrations and their derivatives. The variant without interpolation performs similarly to the MLP proxy, indicating that resampling alone provides little benefit. By contrast, incorporating interpolation before finite differencing yields the lowest rate coefficient error ($\mathcal{L}_k = 0.145$), making it the most effective training strategy among the tested variants.

\subsection{Robustness studies}
In real-world chemistry experiments, measurement frequency is often constrained by instrumentation and cost, leading to sparse observations, sometimes with as few as 10 time points for an entire reaction. To evaluate the robustness of our method under such conditions, we simulate data sparsity by downsampling the AOXID concentration trajectories by factors of 2, 4, and 10, while keeping the total time span fixed. We then assess the accuracy of CRNN pre-training under these scenarios.

As shown in Table~\ref{tab:downsample}, our method remains stable and produces physically plausible estimates even as data sparsity increases. The trajectory loss $\mathcal{L}_y$ and coefficient error increase gradually, but the degradation in performance remains moderate. Even with a 10-fold reduction in observations, the model achieves a reasonable rate coefficient error of 0.45, demonstrating robustness to sparse supervision and applicability to low-resolution experimental datasets.

\begin{table}[t]
\centering
\renewcommand\arraystretch{1.3}
\resizebox{\columnwidth}{!}{
\begin{tabular}{c|clc}
\toprule
 Downsampling rate    &  ODE fitting  &&  Rate coeff. estimate\\ \midrule
w/o downsampling &  $8.9\times10^{-5}$ &&  0.145\\
2  &  $5.3\times10^{-4}$ &&  0.384\\
4  &  $6.4\times10^{-4}$ &&  0.345\\
10 &  $2.0\times10^{-3}$ &&  0.437\\ \bottomrule
\end{tabular}
}
\caption{\textbf{Robustness of CRNN pre-training under increasingly sparse observations.} We downsample the AOXID trajectories and report the ODE fitting and rate coefficient estimation errors in SPIN-ODE-s2.}
\label{tab:downsample}
\end{table}

\section{Conclusion}
\label{conclusion}

We propose a neural network framework for estimating chemical reaction rate coefficients without relying on prior knowledge. To the best of our knowledge, this is the first approach to address stiff chemical ODEs under such conditions. This capability is particularly relevant for real-world systems such as atmospheric chemistry, where rate coefficients vary exponentially and strongly influence reaction rates. Our method follows a three-stage optimisation process: it begins with trajectory fitting of stiff ODEs, proceeds with explicit learning of the full chemical system using CRNN, and concludes with a refinement step that integrates the CRNN with an ODE solver to improve accuracy. We validate the framework on both synthetic datasets and simplified atmospheric scenarios, with results demonstrating state-of-the-art performance in trajectory fitting and rate coefficient estimation. Future work will extend the method to comprehensive chemistry schemes and further strengthen its robustness under real-world conditions.

\begin{ack}
The authors wish to acknowledge CSC – IT Center for Science, Finland, for computational resources. W. P. gratefully acknowledges financial support from the Finnish Cultural Foundation (Päijät-Häme Regional Fund 70251819) and the Onni ja Hilja Tuovinen Foundation Grant.

\end{ack}

\bibliography{main}

\begin{thebibliography}{49}
\providecommand{\natexlab}[1]{#1}
\providecommand{\url}[1]{\texttt{#1}}
\expandafter\ifx\csname urlstyle\endcsname\relax
  \providecommand{\doi}[1]{doi: #1}\else
  \providecommand{\doi}{doi: \begingroup \urlstyle{rm}\Url}\fi

\bibitem[Betancourt et~al.(2023)Betancourt, Li, Kleinert, and Schultz]{air_2}
C.~Betancourt, C.~W.~Y. Li, F.~Kleinert, and M.~G. Schultz.
\newblock Graph machine learning for improved imputation of missing tropospheric ozone data.
\newblock \emph{Environmental Science \& Technology}, 57\penalty0 (46):\penalty0 18246--18258, 2023.
\newblock \doi{10.1021/acs.est.3c05104}.
\newblock PMID: 37661931.

\bibitem[Bi et~al.(2023)Bi, Xie, Zhang, and et~al.]{pangu}
K.~Bi, L.~Xie, H.~Zhang, and et~al.
\newblock Accurate medium-range global weather forecasting with 3d neural networks.
\newblock \emph{Nature}, 619:\penalty0 533--538, 2023.

\bibitem[Bodnar et~al.(2024)Bodnar, Bruinsma, and et~al.]{aurora}
C.~Bodnar, W.~P. Bruinsma, and et~al.
\newblock Aurora: A foundation model of the atmosphere.
\newblock \emph{arXiv}, 2405.13063, 2024.

\bibitem[Butcher(1996)]{rk}
J.~Butcher.
\newblock A history of runge-kutta methods.
\newblock \emph{Applied Numerical Mathematics}, 20\penalty0 (3):\penalty0 247--260, 1996.
\newblock ISSN 0168-9274.
\newblock \doi{https://doi.org/10.1016/0168-9274(95)00108-5}.

\bibitem[Chen et~al.(2018)Chen, Rubanova, Bettencourt, and Duvenaud]{ODE}
T.~Q. Chen, Y.~Rubanova, J.~Bettencourt, and D.~K. Duvenaud.
\newblock Neural ordinary differential equations.
\newblock \emph{Advances in neural information processing systems}, 2018.

\bibitem[Chu et~al.(2021)Chu, Chen, and et~al.]{smog}
B.~Chu, T.~Chen, and et~al.
\newblock Application of smog chambers in atmospheric process studies.
\newblock \emph{National Science Review}, 9\penalty0 (2):\penalty0 nwab103, 06 2021.
\newblock ISSN 2095-5138.
\newblock \doi{10.1093/nsr/nwab103}.

\bibitem[Cuomo et~al.(2022)Cuomo, Di~Cola, and Giampaolo]{pinn}
S.~Cuomo, V.~Di~Cola, and F.~e.~a. Giampaolo.
\newblock Scientific machine learning through physics–informed neural networks: Where we are and what’s next.
\newblock \emph{J Sci Comput}, 92\penalty0 (88), 2022.

\bibitem[Curtiss and Hirschfelder(1952)]{stiff_1}
C.~Curtiss and J.~Hirschfelder.
\newblock Integration of stiff equations.
\newblock \emph{Proceedings of the National Academy of Sciences of the United States of America}, 38\penalty0 (3), 1952.

\bibitem[Damian et~al.(2002)Damian, Sandu, Damian, Potra, and Carmichael]{damian_kinetic_2002}
V.~Damian, A.~Sandu, M.~Damian, F.~Potra, and G.~R. Carmichael.
\newblock The kinetic preprocessor {KPP}-a software environment for solving chemical kinetics.
\newblock \emph{Computers \& Chemical Engineering}, 26\penalty0 (11):\penalty0 1567--1579, Nov. 2002.
\newblock ISSN 00981354.
\newblock \doi{10.1016/S0098-1354(02)00128-X}.

\bibitem[Di et~al.(2020)Di, Amini, and et~al.]{air_4}
Q.~Di, H.~Amini, and et~al.
\newblock Assessing no2 concentration and model uncertainty with high spatiotemporal resolution across the contiguous united states using ensemble model averaging.
\newblock \emph{Environmental Science \& Technology}, 54\penalty0 (3):\penalty0 1372--1384, 2020.
\newblock \doi{10.1021/acs.est.9b03358}.
\newblock PMID: 31851499.

\bibitem[Dupont et~al.(2019)Dupont, Doucet, and Teh]{augmented}
E.~Dupont, A.~Doucet, and Y.~W. Teh.
\newblock Augmented neural odes.
\newblock \emph{Advances in neural information processing systems}, 2019.

\bibitem[Finlay et~al.(2020)Finlay, Jacobsen, Nurbekyan, and Oberman]{ode_1}
C.~Finlay, J.-H. Jacobsen, L.~Nurbekyan, and A.~Oberman.
\newblock How to train your neural ode: the world of jacobian and kinetic regularization.
\newblock In \emph{International conference on machine learning}, pages 3154--3164. PMLR, 2020.

\bibitem[Fronk and Petzold(2024)]{stiff_4}
C.~Fronk and L.~Petzold.
\newblock Training stiff neural ordinary differential equations with implicit single-step methods.
\newblock \emph{Chaos: An Interdisciplinary Journal of Nonlinear Science}, 34\penalty0 (12):\penalty0 123147, 12 2024.

\bibitem[Fronk and Petzold(2025)]{stiff_3}
C.~Fronk and L.~Petzold.
\newblock Training stiff neural ordinary differential equations with explicit exponential integration methods.
\newblock \emph{Chaos: An Interdisciplinary Journal of Nonlinear Science}, 35\penalty0 (3):\penalty0 033154, 03 2025.

\bibitem[Ghosh et~al.(2020)Ghosh, Behl, Dupont, Torr, and Namboodiri]{steer}
A.~Ghosh, H.~Behl, E.~Dupont, P.~Torr, and V.~Namboodiri.
\newblock Steer: Simple temporal regularization for neural odes.
\newblock \emph{Advances in neural information processing systems}, 2020.

\bibitem[Goswami et~al.(2024)Goswami, Jagtap, Babaee, Susi, and Karniadakis]{goswami2024learning}
S.~Goswami, A.~D. Jagtap, H.~Babaee, B.~T. Susi, and G.~E. Karniadakis.
\newblock Learning stiff chemical kinetics using extended deep neural operators.
\newblock \emph{Computer Methods in Applied Mechanics and Engineering}, 419:\penalty0 116674, 2024.

\bibitem[Hou et~al.(2022)Hou, Dai, Song, and et~al.]{air_1}
L.~Hou, Q.~Dai, C.~Song, and et~al.
\newblock Revealing drivers of haze pollution by explainable machine learning.
\newblock \emph{Environmental Science \& Technology Letters}, 9\penalty0 (2):\penalty0 112--119, 2022.
\newblock \doi{10.1021/acs.estlett.1c00865}.

\bibitem[Howard(1979)]{flow_tube}
C.~J. Howard.
\newblock Kinetic measurements using flow tubes.
\newblock \emph{The Journal of Physical Chemistry}, 83:\penalty0 3--9, 1979.

\bibitem[Huang et~al.(2025)Huang, Kandris, and Katsou]{huang_training_2025}
X.~Huang, K.~Kandris, and E.~Katsou.
\newblock Training stiff neural ordinary differential equations in data-driven wastewater process modelling.
\newblock \emph{Journal of Environmental Management}, 373:\penalty0 123870, Jan. 2025.
\newblock ISSN 03014797.
\newblock \doi{10.1016/j.jenvman.2024.123870}.

\bibitem[I. and T.(2023)]{stiff_6}
S.~I. and B.~T.
\newblock Speeding up astrochemical reaction networks with autoencoders and neural odes.
\newblock \emph{Advances in neural information processing systems Workshop}, 2023.

\bibitem[Ingebrand et~al.(2024)Ingebrand, Thorpe, and Topcu]{ode_7}
T.~Ingebrand, A.~Thorpe, and U.~Topcu.
\newblock Zero-shot transfer of neural odes.
\newblock \emph{Advances in Neural Information Processing Systems}, 37:\penalty0 67604--67626, 2024.

\bibitem[Ji and Deng(2021)]{ji_autonomous_2021}
W.~Ji and S.~Deng.
\newblock Autonomous discovery of unknown reaction pathways from data by chemical reaction neural network.
\newblock \emph{The Journal of Physical Chemistry A}, 125\penalty0 (4):\penalty0 1082--1092, Feb. 2021.
\newblock ISSN 1089-5639, 1520-5215.
\newblock \doi{10.1021/acs.jpca.0c09316}.

\bibitem[Ji et~al.(2021)Ji, Qiu, Shi, Pan, and Deng]{ji_stiff-pinn_2021}
W.~Ji, W.~Qiu, Z.~Shi, S.~Pan, and S.~Deng.
\newblock Stiff-pinn: Physics-informed neural network for stiff chemical kinetics.
\newblock \emph{The Journal of Physical Chemistry A}, 125\penalty0 (36):\penalty0 8098--8106, Sept. 2021.
\newblock ISSN 1089-5639, 1520-5215.
\newblock \doi{10.1021/acs.jpca.1c05102}.

\bibitem[Kidger(2022)]{kidger_neural_2022}
P.~Kidger.
\newblock On neural differential equations.
\newblock \emph{arXiv preprint arXiv:2202.02435}, 2022.

\bibitem[Kim et~al.(2021)Kim, Ji, Deng, Ma, and Rackauckas]{kim_stiff_2021}
S.~Kim, W.~Ji, S.~Deng, Y.~Ma, and C.~Rackauckas.
\newblock Stiff neural ordinary differential equations.
\newblock \emph{Chaos: An Interdisciplinary Journal of Nonlinear Science}, 31\penalty0 (9):\penalty0 093122, Sept. 2021.
\newblock ISSN 1054-1500, 1089-7682.
\newblock \doi{10.1063/5.0060697}.

\bibitem[Koenig et~al.(2024)Koenig, Kim, and Deng]{kanode}
B.~C. Koenig, S.~Kim, and S.~Deng.
\newblock Kan-odes: Kolmogorov–arnold network ordinary differential equations for learning dynamical systems and hidden physics.
\newblock \emph{Computer Methods in Applied Mechanics and Engineering}, 432:\penalty0 117397, 2024.
\newblock ISSN 0045-7825.
\newblock \doi{https://doi.org/10.1016/j.cma.2024.117397}.

\bibitem[Koenig et~al.(2025)Koenig, Kim, and Deng]{chemkan}
B.~C. Koenig, S.~Kim, and S.~Deng.
\newblock Chemkans for combustion chemistry modeling and acceleration.
\newblock \emph{arXiv preprint arXiv:2504.12580}, 2025.

\bibitem[Kong et~al.(2022)Kong, Yamashita, Foggo, and Yu]{kong_dynamic_2022}
X.~Kong, K.~Yamashita, B.~Foggo, and N.~Yu.
\newblock Dynamic parameter estimation with physics-based neural ordinary differential equations.
\newblock In \emph{2022 IEEE Power \& Energy Society General Meeting ({PESGM})}, pages 1--5, Denver, CO, USA, July 2022. IEEE.
\newblock ISBN 978-1-6654-0823-3.
\newblock \doi{10.1109/PESGM48719.2022.9916840}.

\bibitem[Lam and et~al.(2023)]{graphcast}
R.~Lam and et~al.
\newblock Learning skillful medium-range global weather forecasting.
\newblock \emph{Science}, 382:\penalty0 1416--1421, 2023.

\bibitem[Li et~al.(2024)Li, Zhang, Zhu, Zhao, Zhang, Duan, and Lin]{ode_6}
X.~Li, J.~Zhang, Q.~Zhu, C.~Zhao, X.~Zhang, X.~Duan, and W.~Lin.
\newblock From fourier to neural odes: Flow matching for modeling complex systems.
\newblock In \emph{Proceedings of the 41st International Conference on Machine Learning}, volume 235 of \emph{Proceedings of Machine Learning Research}, pages 29390--29405, 21--27 Jul 2024.

\bibitem[Liu et~al.(2024)Liu, Wang, Vaidya, Ruehle, Halverson, Solja{\v{c}}i{\'c}, Hou, and Tegmark]{kan}
Z.~Liu, Y.~Wang, S.~Vaidya, F.~Ruehle, J.~Halverson, M.~Solja{\v{c}}i{\'c}, T.~Y. Hou, and M.~Tegmark.
\newblock Kan: Kolmogorov-arnold networks.
\newblock \emph{arXiv preprint arXiv:2404.19756}, 2024.

\bibitem[Liu et~al.(2025)Liu, Clusius, and Boy]{chemode_2}
Z.-S. Liu, P.~Clusius, and M.~Boy.
\newblock Neural network emulator for atmospheric chemical ode.
\newblock \emph{Neural Networks}, 184:\penalty0 107106, 2025.
\newblock ISSN 0893-6080.
\newblock \doi{https://doi.org/10.1016/j.neunet.2024.107106}.

\bibitem[Lu et~al.(2021)Lu, Jin, Pang, Zhang, and Karniadakis]{lu2021learning}
L.~Lu, P.~Jin, G.~Pang, Z.~Zhang, and G.~E. Karniadakis.
\newblock Learning nonlinear operators via deeponet based on the universal approximation theorem of operators.
\newblock \emph{Nature machine intelligence}, 3\penalty0 (3):\penalty0 218--229, 2021.

\bibitem[Massaroli et~al.(2020)Massaroli, Poli, Park, Yamashita, and Asama]{ode_2}
S.~Massaroli, M.~Poli, J.~Park, A.~Yamashita, and H.~Asama.
\newblock Dissecting neural odes.
\newblock \emph{Advances in neural information processing systems}, 2020.

\bibitem[Owoyele and Pal(2022)]{chemode}
O.~Owoyele and P.~Pal.
\newblock Chemnode: A neural ordinary differential equations framework for efficient chemical kinetic solvers.
\newblock \emph{Energy and AI}, 7:\penalty0 100118, 2022.
\newblock ISSN 2666-5468.
\newblock \doi{https://doi.org/10.1016/j.egyai.2021.100118}.

\bibitem[Peng(2025)]{peng_spinode_2025}
W.~Peng.
\newblock Spin-ode: Stiff physics-informed neural ode for chemical reaction rate estimation (v1.0.0), 2025.
\newblock URL \url{https://doi.org/10.5281/zenodo.16949166}.

\bibitem[Peng et~al.(2025)Peng, Liu, and Boy]{peng2025spin}
W.~Peng, Z.-S. Liu, and M.~Boy.
\newblock Spin-ode: Stiff physics-informed neural ode for chemical reaction rate estimation.
\newblock \emph{arXiv preprint arXiv:2505.05625}, 2025.
\newblock Full version of this paper.

\bibitem[Pichelstorfer et~al.(2025)Pichelstorfer, O'Meara, and McFiggans]{pichelstorfer2025theory}
L.~Pichelstorfer, S.~P. O'Meara, and G.~McFiggans.
\newblock A theory-informed, experiment-based constraint on the rate of autoxidation chemistry--an analytical approach.
\newblock \emph{Aerosol Research}, 3\penalty0 (2):\penalty0 417--428, 2025.

\bibitem[Robertson and Walsh(1966)]{robertson1966numerical}
H.~Robertson and J.~Walsh.
\newblock Numerical analysis: an introduction.
\newblock In \emph{Chapitre the Solution of a Set of Reaction Rate Equations}. Academic Press, 1966.

\bibitem[Roesch et~al.(2021)Roesch, Rackauckas, and Stumpf]{roesch_collocation_2021}
E.~Roesch, C.~Rackauckas, and M.~P.~H. Stumpf.
\newblock Collocation based training of neural ordinary differential equations.
\newblock \emph{Statistical Applications in Genetics and Molecular Biology}, 20\penalty0 (2):\penalty0 37--49, July 2021.
\newblock ISSN 1544-6115, 2194-6302.
\newblock \doi{10.1515/sagmb-2020-0025}.

\bibitem[Salunkhe et~al.(2022)Salunkhe, Deighan, DesJardin, and Chandola]{chemtab}
A.~Salunkhe, D.~Deighan, P.~E. DesJardin, and V.~Chandola.
\newblock Chemtab: A physics guided chemistry modeling framework.
\newblock In \emph{International Conference on Computational Science}, pages 75--88. Springer, 2022.

\bibitem[Seiler et~al.(2025)Seiler, Lei, and Protopapas]{stiff_5}
E.~Seiler, W.~Lei, and P.~Protopapas.
\newblock Stiff transfer learning for physics-informed neural networks.
\newblock \emph{arXiv preprint arXiv:2501.17281}, 2025.

\bibitem[Su et~al.(2023)Su, Ji, An, Ren, Deng, and Law]{su_kinetics_2023}
X.~Su, W.~Ji, J.~An, Z.~Ren, S.~Deng, and C.~K. Law.
\newblock Kinetics parameter optimization of hydrocarbon fuels via neural ordinary differential equations.
\newblock \emph{Combustion and Flame}, 251:\penalty0 112732, May 2023.
\newblock ISSN 00102180.
\newblock \doi{10.1016/j.combustflame.2023.112732}.

\bibitem[Th\"oni et~al.(2025)Th\"oni, Robinson, Bachrach, Huck, and Kachman]{reactionnet}
A.~C. Th\"oni, W.~E. Robinson, Y.~Bachrach, W.~T. Huck, and T.~Kachman.
\newblock Modeling chemical reaction networks using neural ordinary differential equations.
\newblock \emph{Journal of Chemical Information and Modeling}, 2025.

\bibitem[Verwer(1994)]{pollu}
J.~G. Verwer.
\newblock Gauss–seidel iteration for stiff odes from chemical kinetics.
\newblock \emph{SIAM Journal on Scientific Computing}, 15\penalty0 (5):\penalty0 1243--1250, Sept. 1994.
\newblock ISSN 1064-8275, 1095-7197.
\newblock \doi{10.1137/0915076}.

\bibitem[Yildiz et~al.(2019)Yildiz, Heinonen, and Lahdesmaki]{ode_4}
C.~Yildiz, M.~Heinonen, and H.~Lahdesmaki.
\newblock Ode2vae: Deep generative second order odes with bayesian neural networks.
\newblock \emph{Advances in Neural Information Processing Systems}, 32, 2019.

\bibitem[Yulia~Rubanova(2019)]{ode_3}
D.~D. Yulia~Rubanova, Ricky T. Q.~Chen.
\newblock Latent odes for irregularly-sampled time series.
\newblock \emph{Advances in neural information processing systems}, 2019.

\bibitem[Zhu et~al.(2022)Zhu, Laughner, and Cohen]{air_3}
Q.~Zhu, J.~L. Laughner, and R.~C. Cohen.
\newblock Combining machine learning and satellite observations to predict spatial and temporal variation of near surface oh in north american cities.
\newblock \emph{Environmental Science \& Technology}, 56\penalty0 (11):\penalty0 7362--7371, 2022.
\newblock \doi{10.1021/acs.est.1c05636}.
\newblock PMID: 35302754.

\bibitem[Zou et~al.(2024)Zou, Levine, Zaharieva, Johari, and Fox]{ode_5}
B.~Zou, E.~Levine, P.~Zaharieva, R.~Johari, and E.~Fox.
\newblock Hybrid$^2$ neural {ODE} causal modeling and an application to glycemic response.
\newblock In \emph{Proceedings of the 41st International Conference on Machine Learning}, volume 235 of \emph{Proceedings of Machine Learning Research}, pages 62934--62963. PMLR, 21--27 Jul 2024.

\end{thebibliography}

\end{document}


\begin{frontmatter}

\paperid{1405}

\title{Supplementary: SPIN-ODE: Stiff Physics-Informed Neural ODE for Chemical Reaction Rate Estimation}

\author[A,C]{\fnms{Wenqing}~\snm{Peng}\orcid{0009-0008-4177-7576}\thanks{Corresponding Author. Email: wenqing.peng@helsinki.fi.}}
\author[B,C]{\fnms{Zhi-Song}~\snm{Liu}\orcid{0000-0003-4507-3097}}
\author[A,B,C]{\fnms{Michael}~\snm{Boy}\orcid{0000-0002-8107-4524}} 

\address[A]{Institute for Atmospheric and Earth Systems Research, University of Helsinki}
\address[B]{Department of Computational Engineering, Lappeenranta-Lahti University of Technology LUT}
\address[C]{Atmospheric Modelling Centre-Lahti}

\end{frontmatter}

\section{Reaction schemes and initial values}

\subsection{POLLU: Air pollution reaction scheme~\cite{pollu}}

\begin{table*}[ht]
\centering
\renewcommand\arraystretch{1}
\resizebox{0.8\textwidth}{!}{
\begin{tabular}{llcc}
\toprule
 \multicolumn{2}{c}{\textbf{Reaction}}& \textbf{Rate Coeff.}& \textbf{Predicted Rate Coeff.}\\ \hline
 01&NO2 = NO + O3P & 0.350E+00 & 0.249E+00 \\
 02&NO + O3 = NO2 & 0.266E+02 & 0.266E+02 \\
 03&HO2 + NO = NO2 + OH & 0.120E+05 & 0.116E+05 \\
 04&HCHO = HO2 + HO2 + CO & 0.860E-03 & 0.842E-03 \\
 05&HCHO = CO & 0.820E-03 & 0.255E-02 \\
 06&HCHO + OH = HO2 + CO & 0.150E+05 & 0.138E+05 \\
 07&ALD = MEO2 + HO2 + CO & 0.130E-03 & 0.334E-03 \\
 08&ALD + OH = C2O3 & 0.240E+05 & 0.300E+05 \\
 09&C2O3 + NO = NO2 + MEO2 + CO2 & 0.165E+05 & 0.205E+05 \\
 10&C2O3 + NO2 = PAN & 0.900E+04 & 0.113E+05 \\
 11&PAN = C2O3 + NO2 & 0.220E-01 & 0.105E-01 \\
 12&MEO2 + NO = CH3O + NO2 & 0.120E+05 & 0.156E+05 \\
 13&CH3O = HCHO + HO2 & 0.188E+01 & 0.321E+01 \\
 14&NO2 + OH = HNO3 & 0.163E+05 & 0.210E+05 \\
 15&O3P = O3 & 0.480E+07 & - \\
 16&O3 = O1D & 0.350E-03 & - \\
 17&O3 = O3P & 0.175E-01 & - \\
 18&O1D = OH + OH & 0.100E+09 & - \\
 19&O1D = O3P & 0.444E+12 & - \\
 20&SO2 + OH = SO4 + HO2 & 0.124E+04 & 0.150E+04 \\
 21&NO3 = NO & 0.210E+01 & 0.741E+01 \\
 22&NO3 = NO2 + O3P & 0.578E+01 & 0.609E+01 \\
 23&NO2 + O3 = NO3 & 0.474E-01 & 0.054E+00 \\
 24&NO3 + NO2 = N2O5 & 0.178E+04 & 0.148E+04 \\
 25&N2O5 = NO3 + NO2 & 0.312E+01 & 0.071E+00 \\
\bottomrule
\end{tabular}
}
\caption{POLLU reaction scheme with ground truth and predicted rate coefficients}
\label{tab:POLLU}
\end{table*}

Reaction pathways, rate coefficients, and our estimations for the POLLU problem are listed in Table~\ref{tab:POLLU}. Initial concentrations for NO, O3, HCHO, CO, ALD, and SO2 are 0.2, 0.04, 0.1, 0.3, 0.01, and 0.007, respectively. Other concentrations are initialised as 0.

\subsection{AOXID: Toy autoxidation reaction scheme~\cite{pichelstorfer2025theory}}

\begin{table*}[ht]
\renewcommand\arraystretch{1}
\resizebox{0.9\textwidth}{!}{
\centering
\begin{tabular}{llcc}
\hline
 \multicolumn{2}{c}{\textbf{Reaction}}& \textbf{Rate Coeff.} & \textbf{Predicted Rate Coeff.} \\ \hline
 01&TOY + OH = T$\_$RO2$\_$O3                    & 1.000E-11                       & 1.002E-11                           \\
 02&TOY + O3 = T$\_$RO2$\_$O2 + OH               & 1.000E-15                       & 1.001E-15                           \\
 03&T$\_$RO2$\_$O2 = T$\_$RO2$\_$O4              & 1.000E+00                       & 1.005E+00                           \\
 04&T$\_$RO2$\_$O3 = T$\_$RO2$\_$O5              & 1.000E+00                       & 1.002E+00                           \\
 05&T$\_$RO2$\_$O2 + NO = T$\_$O0$\_$NO3         & 4.097E-12                       & 4.094E-12                           \\
 06&T$\_$RO2$\_$O4 + NO = T$\_$O2$\_$NO3         & 4.097E-12                       & 4.100E-12                           \\
 07&T$\_$RO2$\_$O3 + NO = T$\_$O1$\_$NO3         & 4.097E-12                       & 4.111E-12                           \\
 08&T$\_$RO2$\_$O5 + NO = T$\_$O3$\_$NO3         & 4.097E-12                       & 4.108E-12                           \\
 09&T$\_$RO2$\_$O2 + NO = T$\_$RO$\_$O1 + NO2    & 6.146E-12                       & 1.571E-12                           \\
 10&T$\_$RO2$\_$O4 + NO = T$\_$RO$\_$O3 + NO2    & 6.146E-12                       & 8.487E-12                           \\
 11&T$\_$RO2$\_$O3 + NO = T$\_$RO$\_$O2 + NO2    & 6.146E-12                       & 8.908E-12                           \\
 12&T$\_$RO2$\_$O5 + NO = T$\_$RO$\_$O4 + NO2    & 6.146E-12                       & 6.148E-12                           \\
 13&T$\_$RO$\_$O4 = fragdummy                    & 1.000E+06                       & 9.983E+05                           \\
 14&T$\_$RO$\_$O2 = T$\_$RO2$\_$O4               & 1.000E+06                       & 1.443E+06                           \\
 15&T$\_$RO$\_$O1 = T$\_$RO2$\_$O3               & 1.000E+06                       & 3.406E+05                           \\
 16&T$\_$RO$\_$O3 = T$\_$RO2$\_$O5               & 1.000E+06                       & 1.384E+06                           \\
 17&T$\_$RO2$\_$O2 = Tuni$\_$O1$\_$O             & 1.000E+00                       & 1.007E+00                           \\
 18&T$\_$RO2$\_$O4 = Tuni$\_$O3$\_$O             & 1.000E+00                       & 1.002E+00                           \\
 19&T$\_$RO2$\_$O3 = Tuni$\_$O2$\_$O             & 1.000E+00                       & 1.003E+00                           \\
 20&T$\_$RO2$\_$O5 = Tuni$\_$O4$\_$O             & 1.000E+00                       & 1.002E+00                           \\
 21&T$\_$RO2$\_$O2 + HO2 = T$\_$O0$\_$OOH        & 3.230E-11                       & 3.242E-11                           \\
 22&T$\_$RO2$\_$O4 + HO2 = T$\_$O2$\_$OOH        & 3.230E-11                       & 3.228E-11                           \\
 23&T$\_$RO2$\_$O3 + HO2 = T$\_$O1$\_$OOH        & 3.230E-11                       & 3.237E-11                           \\
 24&T$\_$RO2$\_$O5 + HO2 = T$\_$O3$\_$OOH        & 3.230E-11                       & 3.232E-11                           \\
 25&T$\_$RO2$\_$O2 + HO2 = T$\_$RO$\_$O1 + OH    & 3.589E-12                       & 3.740E-12                           \\
 26&T$\_$RO2$\_$O4 + HO2 = T$\_$RO$\_$O3 + OH    & 3.589E-12                       & 3.723E-12                           \\
 27&T$\_$RO2$\_$O3 + HO2 = T$\_$RO$\_$O2 + OH    & 3.589E-12                       & 3.786E-12                           \\
 28&T$\_$RO2$\_$O5 + HO2 = T$\_$RO$\_$O4 + OH    & 3.589E-12                       & 3.053E-12                           \\
 29&T$\_$RO2$\_$O4 = T$\_$O1$\_$2OH              & RO2$\times$3.000E-14& RO2$\times$3.016E-14                           \\
 30&T$\_$RO2$\_$O3 = T$\_$O0$\_$2OH              & RO2$\times$3.000E-14& RO2$\times$3.016E-14                           \\
 31&T$\_$RO2$\_$O5 = T$\_$O2$\_$2OH              & RO2$\times$3.000E-14& RO2$\times$3.017E-14                           \\
 32&T$\_$RO2$\_$O2 = T$\_$RO$\_$O1               & RO2$\times$3.000E-14& RO2$\times$2.212E-14                           \\
 33&T$\_$RO2$\_$O4 = T$\_$RO$\_$O3               & RO2$\times$3.000E-14& RO2$\times$4.275E-14                           \\
 34&T$\_$RO2$\_$O3 = T$\_$RO$\_$O2               & RO2$\times$3.000E-14& RO2$\times$4.438E-14                           \\
 35&T$\_$RO2$\_$O5 = T$\_$RO$\_$O4               & RO2$\times$3.000E-14& RO2$\times$3.050E-14                           \\
 36&T$\_$RO2$\_$O2 = T$\_$O0$\_$O                & RO2$\times$4.000E-14& RO2$\times$4.017E-14                           \\
 37&T$\_$RO2$\_$O4 = T$\_$O2$\_$O                & RO2$\times$4.000E-14& RO2$\times$4.022E-14                           \\
 38&T$\_$RO2$\_$O3 = T$\_$O1$\_$O                & RO2$\times$4.000E-14& RO2$\times$4.023E-14                           \\
 39&T$\_$RO2$\_$O5 = T$\_$O3$\_$O                & RO2$\times$4.000E-14& RO2$\times$4.023E-14                           \\
 40&T$\_$RO2$\_$O2 + T$\_$RO2$\_$O2 = TO1$\_$TO1 & 1.000E-13                       & 1.005E-13                           \\
 41&T$\_$RO2$\_$O2 + T$\_$RO2$\_$O4 = TO1$\_$TO3 & 1.000E-13                       & 1.001E-13                           \\
 42&T$\_$RO2$\_$O2 + T$\_$RO2$\_$O3 = TO1$\_$TO2 & 1.000E-13                       & 1.002E-13                           \\
 43&T$\_$RO2$\_$O2 + T$\_$RO2$\_$O5 = TO1$\_$TO4 & 1.000E-13                       & 1.002E-13                           \\
 44&T$\_$RO2$\_$O4 + T$\_$RO2$\_$O4 = TO3$\_$TO3 & 1.000E-13                       & 1.002E-13                           \\
 45&T$\_$RO2$\_$O4 + T$\_$RO2$\_$O3 = TO3$\_$TO2 & 1.000E-13                       & 1.003E-13                           \\
 46&T$\_$RO2$\_$O4 + T$\_$RO2$\_$O5 = TO3$\_$TO4 & 1.000E-13                       & 1.003E-13                           \\
 47&T$\_$RO2$\_$O3 + T$\_$RO2$\_$O3 = TO2$\_$TO2 & 1.000E-13                       & 1.002E-13                           \\
 48&T$\_$RO2$\_$O3 + T$\_$RO2$\_$O5 = TO2$\_$TO4 & 1.000E-13                       & 1.002E-13                           \\
 49&T$\_$RO2$\_$O5 + T$\_$RO2$\_$O5 = TO4$\_$TO4 & 1.000E-13                       & 1.002E-13                           \\ 
\hline
\end{tabular}
}
\caption{AOXID reaction scheme with ground truth and predicted rate coefficients}
\label{tab: AOXID}
\end{table*}

Reaction pathways, rate coefficients, and our estimations for the AOXID problem are listed in Table~\ref{tab: AOXID}, where RO2 denotes the summed concentration of the species T\_RO2\_O2, T\_RO2\_O3, T\_RO2\_O4, and T\_RO2\_O5. Initial concentrations for species TOY, OH, O3, HO2, and NO are 3.10E+11, 1.90E+5, 1.80E+12, 4.30E+7 and 1.10E+8, respectively. Other concentrations are initialised as 0.

\section{Training convergence}
\begin{figure*}
    \centering
    \includegraphics[width=1\linewidth]{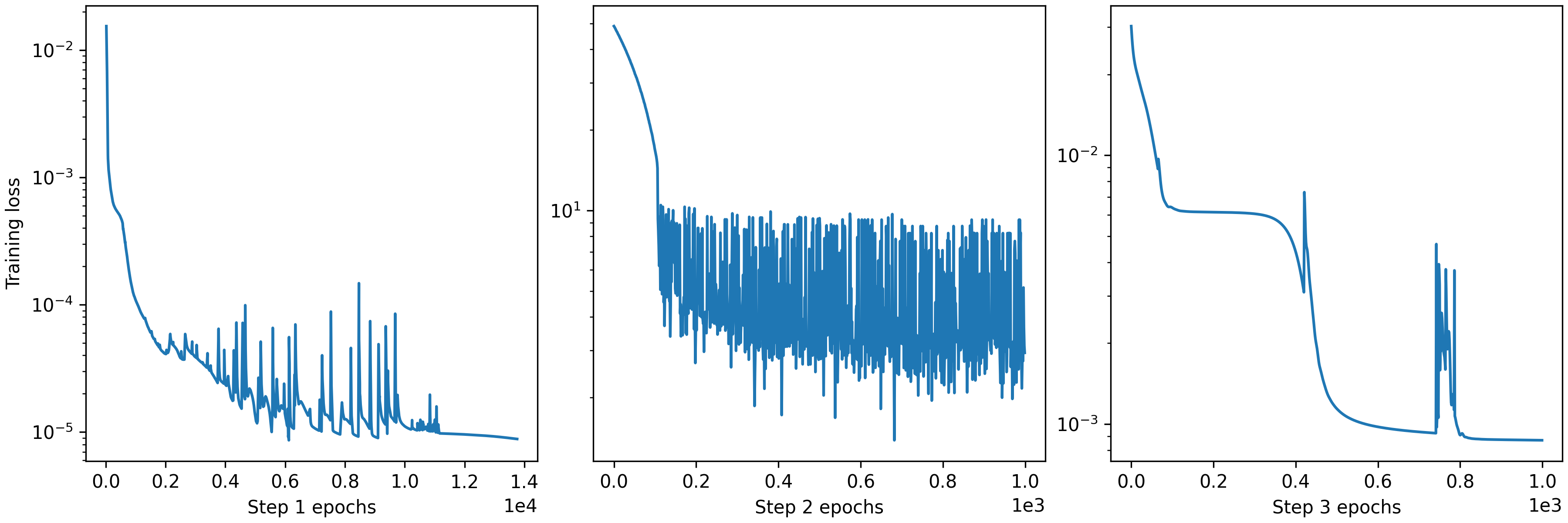}
    \caption{Convergence of SPIN-ODE on the POLLU dataset: step 1 (trajectory fitting), step 2 (CRNN pre-training), and step 3 (CRNN fine-tuning). Loss is shown on a logarithmic scale.}
    \label{fig:convergence}
\end{figure*}

For POLLU problem, training convergence of each stage is shown in Figure \ref{fig:convergence}.

\section{Complete concentration trajectory fitting for POLLU and AOXID problems.}

\begin{figure*}[ht]
    \centering
    \includegraphics[width=0.9\linewidth]{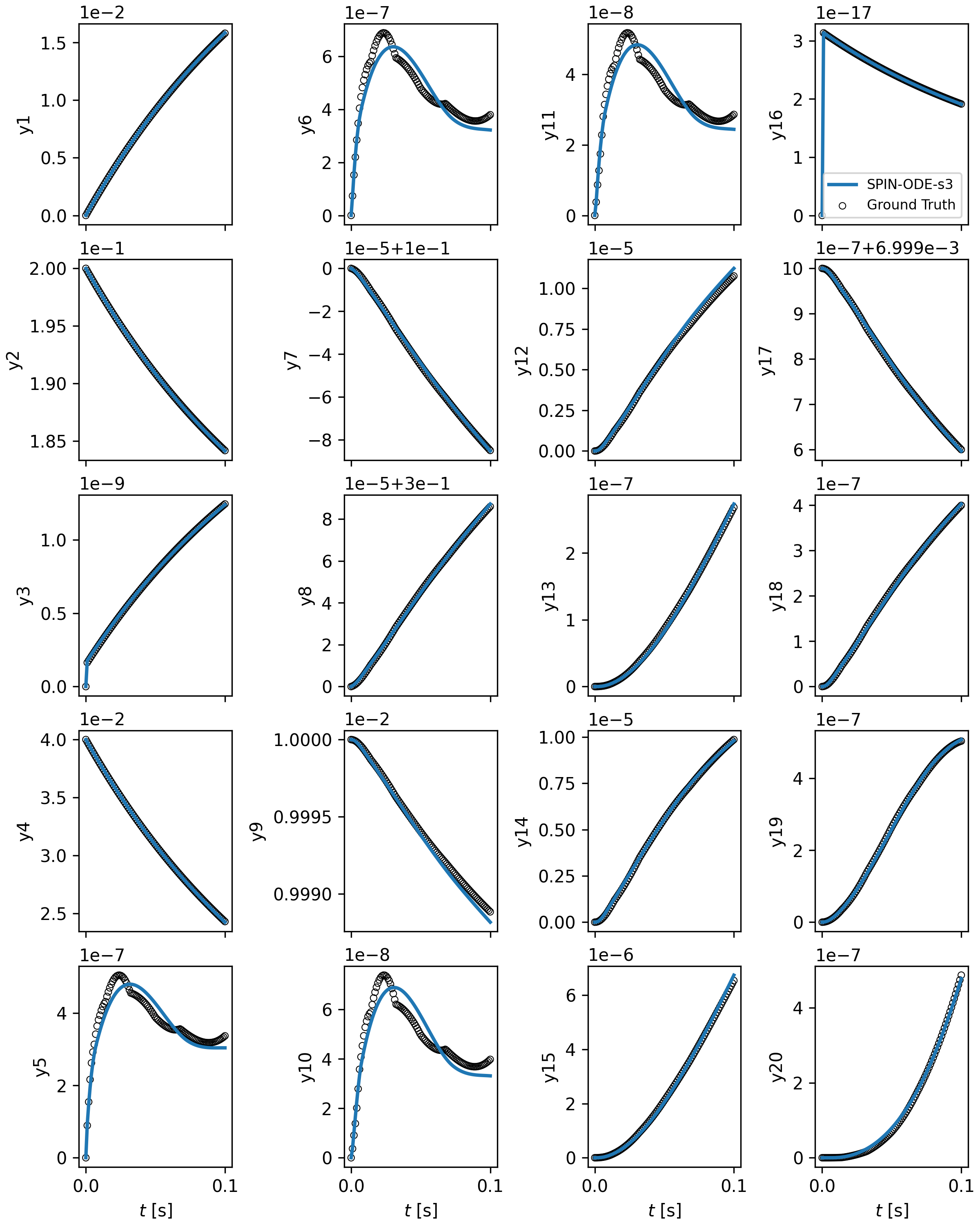}
    \caption{Concentration trajectory fitting for the POLLU problem.}
    \label{fig:fit_y_polly}
\end{figure*}

\begin{figure*}[ht]
    \centering
    \includegraphics[width=0.9\linewidth]{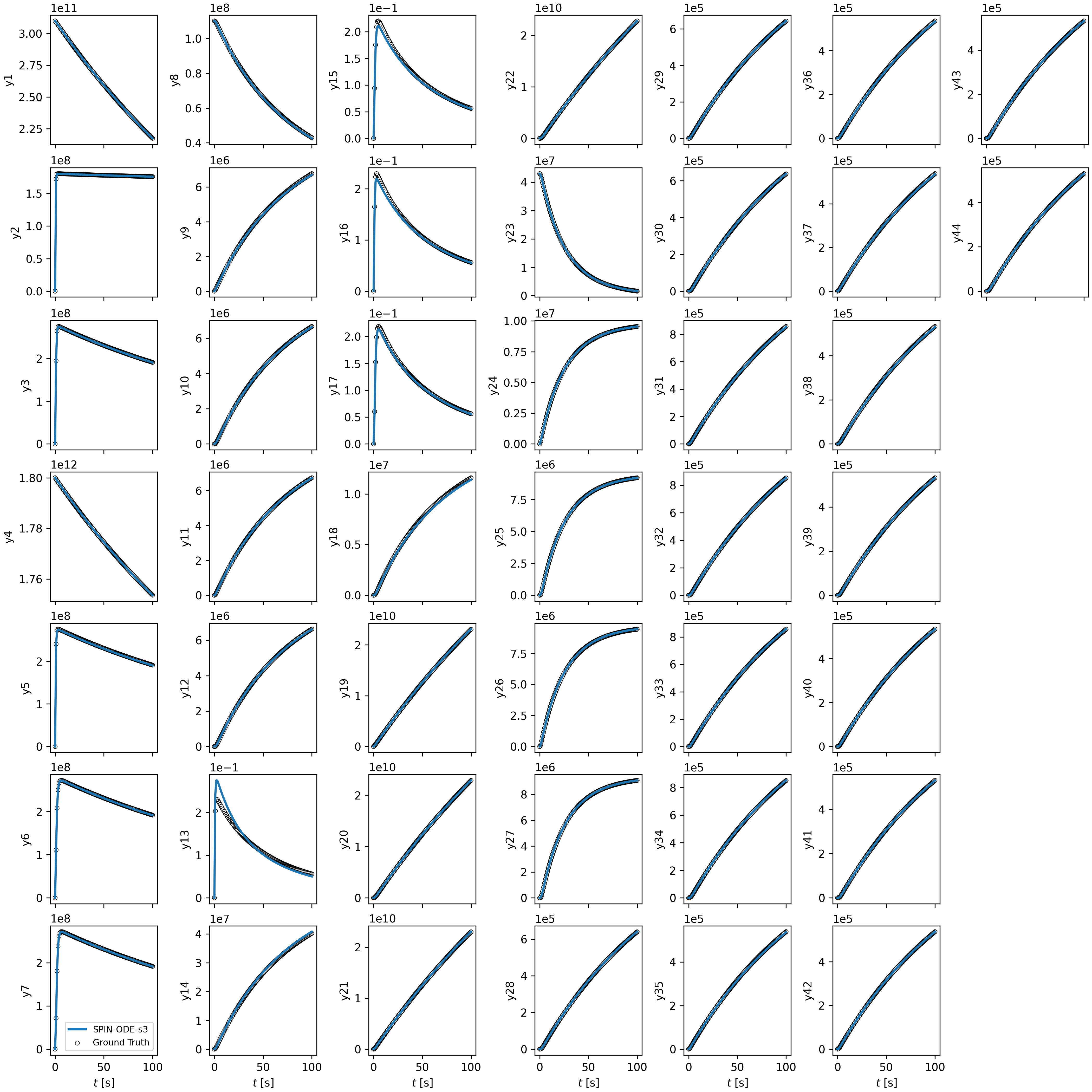}
    \caption{Concentration trajectory fitting for the AOXID problem.}
    \label{fig:fit_y_aoxid}
\end{figure*}

The complete concentration trajectory fittings for POLLU (20 chemical reactions) and AOXID (44 chemical reactions) problems are shown in Figure~\ref{fig:fit_y_polly} and Figure~\ref{fig:fit_y_aoxid}.

